\documentclass{article}
\usepackage{ijcai23}

\usepackage{times}
\usepackage{soul}
\usepackage{url}
\usepackage[hidelinks]{hyperref}
\usepackage[utf8]{inputenc}
\usepackage[small,center]{caption}
\usepackage{graphicx}
\usepackage{amsmath}
\usepackage{amsthm}
\usepackage{booktabs}
\usepackage{algorithm}
\usepackage{algorithmic}
\usepackage[switch]{lineno}

\usepackage{amsfonts}

\newcommand{\A}{{\cal A}}

\newcommand{\M}{{\cal M}}

\newcommand{\zug}[1]{\langle #1  \rangle}
\newcommand{\stam}[1]{}
\newcommand{\set}[1]{\{ #1  \}}

\newcommand{\Rat}{\mathbb{Q}}

\newtheorem*{theorem*}{Theorem}
\newtheorem*{corollary*}{Corollary}
\newtheorem{remark}{Remark}

\newtheorem{example}{Example}
\newtheorem{theorem}{Theorem}

\title{ASQ-IT: Interactive Explanations for Reinforcement-Learning Agents}

\author{
Yotam Amitai$^1$
\and
Guy Avni$^2$\and
Ofra Amir$^1$
\affiliations
$^1$Faculty of Data \& Decision Science, Technion - I.I.T\\
$^2$Faculty of Computer Science, University of Haifa\\
\emails
yotama@campus.technion.ac.il,
gavni@cs.haifa.ac.il,
oamir@technion.ac.il
}

\begin{document}

\maketitle

\begin{abstract}
As reinforcement learning methods increasingly amass accomplishments, the need for comprehending their solutions becomes more crucial.  
Most explainable reinforcement learning (XRL) methods generate a static explanation depicting their developers' intuition of what should be explained and how. In contrast, literature from the social sciences proposes that meaningful explanations are structured as a dialog between the explainer and the explainee, suggesting a more active role for the user and her communication with the agent.
In this paper, we present ASQ-IT -- an interactive tool that presents video clips of the agent acting in its environment based on queries given by the user that describe temporal properties of behaviors of interest. Our approach is based on formal methods: queries in ASQ-IT's user interface map to a fragment of Linear Temporal Logic over finite traces (LTLf), which we developed, and our algorithm for query processing is based on automata theory. 
User studies show that end-users can understand and formulate queries in ASQ-IT, and that using ASQ-IT assists users in identifying faulty agent behaviors.
\end{abstract}

\section{Introduction}
Reinforcement Learning (RL) has shown impressive success in recent years; e.g., mastering Go or achieving human-level performance in Atari games \cite{silver2016mastering,mnih2015human}.
However, current training techniques are complex and rely on implicit goals and indirect feature representations, and thus largely produce black-box agents. In order for such trained agents to be successfully deployed, in particular in safety-critical domains such as healthcare, it is crucial for them to be trustworthy; namely, both developers and users need to understand, predict and assess agents' behavior. This need has led to an abundance of ``explainable RL'' (XRL) methods~\cite{dazeley2021explainable} designed to elucidate black-box agents. 

Existing approaches to XRL are for the most part static. That is, they provide the user with some information about the agent's decision-making. For example, local explanations might show saliency maps depicting the agent's attention, a causal explanation, or an explanation of the reward function. Global explanations might describe the agent's policy by presenting a simplified representation (e.g., a decision tree), or by demonstrating the behavior of the agent through policy summaries. Common to all of these approaches is that the users do not have a way to interact with the provided information or pose questions that they are interested in. 

Following the literature on explanations from the social sciences~\cite{miller2018explanation}, we aim to develop {\em interactive XRL methods} that allow for a dialog between the explainer (system) and the explainee (user): the user repeatedly poses queries for the system to answer. 
Interactive explanations have recently been identified as a significant future direction for system intelligibility and enhancing user engagement~\cite{abdul2018trends}.
Increasing evidence also points towards interaction and exploration as means to reduce over-reliance on AI recommendations, which occurs even when explanations are provided \cite{buccinca2021trust}.


In this work, we develop ``ASQ-IT'', an interactive XRL tool that aims to assist users to comprehend an agent in a global manner. Inspired by policy summarization approaches that demonstrate the behavior of an agent in selected world-states~\cite{amir2019summarizing}, our tool generates clips of the agent interacting with its environment. The user controls which clips will be presented by formulating queries that specify properties of clips of interest. The interaction with the tool resembles a dialogue: the user enters a query, and receives clips that match it; the user can then refine her query, and the process continues. For instance, for a self-driving car agent, the user might formulate a query for examining the agent's ability to switch lanes by specifying a start lane and end lane and our tool will output clips of the agent making this transition. 

The main challenge in developing an interactive tool is the interaction with human users (especially laypeople). Indeed, unless constrained, study participants pose vague and informal queries that are hard for a tool to process. A tool's interface must strike the right balance between expressivity and usability. We address these challenges as follows. {\em i)} We develop a simple logic that can express common properties of clips. Note that clips are sequential, thus our logic must reason about temporal behaviors. An established logic to reason about such properties is Linear Temporal Logic \cite{Pnu77}, and our logic relies on its finite counterpart called LTLf \cite{de2014reasoning}. \emph{ii)} Laypeople cannot be expected to produce logic formulas, thus we develop a simple user interface that maps directly to our logic. \emph{iii)} We assume access to a library of agent execution traces.
We develop an efficient automata-based algorithm to search this library for clips that answer a user’s query. 

Our paper makes the following contributions:
It introduces ASQ-IT, an \textbf{A}gent \textbf{S}ystem \textbf{Q}ueries \textbf{I}nteractive \textbf{T}ool that enables users to describe and generate queries towards an agent and receive answers as explanations-through-demonstration of their behavior.
We present results from two user studies. The first user study shows that laypeople, with no training in logic or RL, are able to comprehend and generate meaningful queries to ASQ-IT. The second study shows that users with some AI background can identify faulty agent behaviors using ASQ-IT and that using ASQ-IT led to improved performance compared to a static policy summary baseline.

\section{Related Work}
This work relates to two 
main areas of research, which we discuss in this section: (1) explanations in sequential decision-making settings and (2) interactive explanations.

\paragraph{Explanations in sequential decision-making settings} In this paper, we focus on the problem of explaining the behavior of agents operating in sequential decision-making settings. Work in this area is typically concerned with explaining policies learned through Reinforcement Learning.

RL explanation methods can be roughly divided into two classes. \emph{Local} explanations focus on explaining specific agent decisions~\cite{krarup2019model,khan2011automatically,hayes2017improving,booth2019evaluating,anderson2020mental}, e.g., by showing what information a game-playing agent attends to in a specific game state~\cite{greydanus2017visualizing}, or generating causal explanations~\cite{madumal2020explainable}. 
In contrast, \emph{global} explanations aim to convey the agent's policy rather than explain particular decisions. One approach to global explanations is to generate a proxy model of the policy that is more interpretable, e.g., through policy graphs~\cite{topin2019generation} or decision trees approximating the policy~\cite{coppens2019distilling}. 
In this paper, we utilize the idea of extracting demonstrations of agent behavior as a global explanation~\cite{amir2019summarizing} to answer queries posed by users, such that they can interactively explore the agent's policy and its characteristics.

\paragraph{Interactive explanations}
Some early works on decision-support systems provided users with interactive explanation methods. For example, MYCIN~\cite{davis1977production}, a system for clinical decision-support, allowed its users to pose ``why'' and ``how'' questions and responded by revealing the rules that led to a particular inference. Such explanations are more difficult to provide in current systems that do not use a logic-based representation. Few works in interpretable machine learning also designed interactive explanations for supervised learning models. For instance, TCAV is a method that enables users to test whether the model relies on a user-determined concept in its decision-making~\cite{kim2018interpretability}. Recently, this approach has been applied to analyzing the chess knowledge of AlphaZero~\cite{mcgrath2021acquisition}. 
Interactive XRL has been flagged as a promising research direction in interactive RL research \cite{arzate2020survey}.
Most closely related to the problem we discuss are the works of \citeauthor{hayes2017improving} \shortcite{hayes2017improving},  \citeauthor{rupprecht2019finding}\shortcite{rupprecht2019finding} and \citeauthor{cruz2021interactive} \shortcite{cruz2021interactive}, each of which introduce systems to help their users debug agent behavior through interactive interfaces. Both works shape the user's interaction through a limited set of action-related questions such as ``when a particular action will be taken?'' or ``why wasn't an alternative action chosen?'', while our focus seeks to bestow more freedom for expressivity and exploration.


\section{ASQ-IT}
In this section we describe the implementation of ASQ-IT. This includes both the backend algorithmic approach, as well as the front-end user interface design.

\begin{figure}[ht]
    \frame{\includegraphics[width=1\columnwidth]{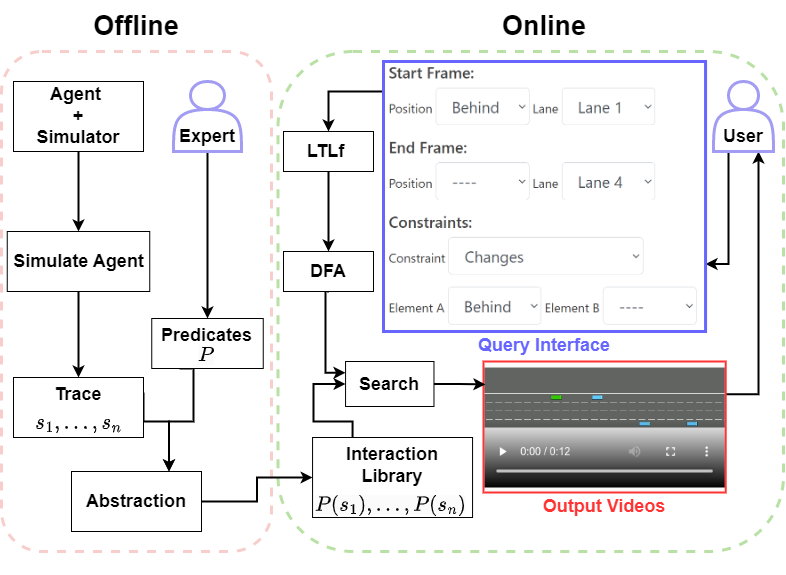}}\\
    \caption{ASQ-IT Process Flow Diagram. \\Example output video: \url{https://bit.ly/3GJV394}}
	\label{fig: tool}
	\vspace{-0.2cm}
\end{figure}

\paragraph{Tool usage, an illustration}  
The users' main interaction point with ASQ-IT is through the \emph{Query Interface} (Fig.~\ref{fig: tool}) where they define scenarios and behaviors they wish to observe in the agent's interaction with the environment. The front-end is based on drop-down menus that depend on predefined predicates given by a domain expert. In the back-end, the user's entries are translated into a formal specification that describes the set of traces that the user is interested in. Based on this specification, the interaction-library database is searched for video clips that answer the user's query. These video clips are then presented to the user.

\emph{Running example: The Highway domain.} The domain consists of a multiple-lane highway in which the agent controls a car depicted by a green rectangle. Other uncontrollable cars are depicted as blue rectangles. Cars can accelerate, decelerate, and change lanes (numbered from top to bottom). We consider various agent goals; for example, a combination of not crashing, driving fast, driving in the right lane, etc.

The following sections describe the building blocks required for constructing and running ASQ-IT.

\subsection{Offline: Obtaining a Database of Clips}
\label{sec:abstractMDP}
We assume access to an agent that operates in an MDP setting. Formally, an MDP is a tuple $\M = \langle S, A, Tr, R \rangle$, where $S$ is a set of states, $A$ is a set of actions, $R: S \rightarrow \Rat$ is a reward function, and $Tr: S\times A \times S \rightarrow [0,1]$ is a probabilistic transition function. An agent is a {\em policy} $\pi$, which is a function $\pi: S \rightarrow A$. We do not assume any knowledge of $Tr$ or $R$. We assume that $\pi$ is given, e.g., trained using RL, and we assume that we have the ability to simulate $\pi$ on $\M$, e.g., using a simulator. This provides a collection of {\em traces}, where each trace is a sequence $s_1, s_2, \ldots, s_n$ of states, i.e., $s_i \in S$, for $1 \leq i \leq n$. For ease of presentation, we assume that the agent is simulated once to produce one trace. In practice, we collect numerous traces and concatenate them -- the more traces collected, the more clips our tool will be able to retrieve in response to user queries. 
 
The goal of our tool is to present to the user a sub-trace $s_k,\ldots, s_\ell$, for $1 \leq k < \ell \leq n$, that the user is interested in. We found that it is infeasible for users to specify a desired behavior directly on the {\em concrete} states. Instead, users' queries are formulated on a predefined collection of {\em predicates} $P$ that are chosen by a domain expert. Each predicate $p \in P$ is a function $p: S \rightarrow \set{\texttt{True},\texttt{False}}$ denoting whether some attribute exists in a state. For example, in the highway domain, the predicate \texttt{lane-1} returns \texttt{True} iff the agent (green car) is in Lane~1 at a given state and the predicate \texttt{behind} returns \texttt{True} iff the agent is driving behind some blue car. For a concrete state $s \in S$, we denote by $P(s)$, the {\em abstract state} that consists of the subset of predicates that hold in $s$, thus $P(s) = \set{p \in P: p(s) = \texttt{True}}$. For example, for $P = \set{\texttt{lane-1}, \texttt{lane-2}, \texttt{behind}}$ and $P(s) = \set{\texttt{lane-1}, \texttt{behind}}$, necessarily at state $s$, the agent is traveling in Lane~$1$ {\em and} behind a blue car. 

To summarize, offline, we simulate the agent $\pi$ on $\M$ to collect a concrete trace $s_1, \ldots, s_n$. A domain expert provides a collection of predicates $P$. Our database consists of both the concrete and abstract trace $P(s_1), \ldots, P(s_n)$. Queries will be processed on the abstract trace, where an answer to a query is $P(s_k),\ldots, P(s_\ell)$, and the corresponding concrete trace $s_k, \ldots, s_\ell$ is presented to the user.

\subsection{Front-End: Query Language and Interface}
One key novelty of ASQ-IT is that it allows users to query for traces that they are interested in. In this section, we describe the formal basis on which our query language is based. We start by surveying the necessary background on Linear Temporal Logic on Finite Traces (LTLf).

\subsubsection{Background: Linear Temporal Logic on Finite Traces}
\label{sec:LTLf}
An LTLf formula $\varphi$ over a collection of predicates $P$ specifies a set of traces; namely, the set of traces that satisfy $\varphi$. We thus think of $\varphi$ as a query. That is, by providing $\varphi$, the user states that she is interested in viewing traces that satisfy $\varphi$.

\begin{example}
\normalfont
We illustrate the syntax and semantics of LTLf. Let $P = \set{\texttt{lane-1}, \texttt{behind}}$. 
\begin{itemize}
\item The formula $X \ \texttt{lane-1}$ (read ``next Lane~$1$'') specifies traces in which the agent is driving in Lane~$1$ in the second position of the trace. No restrictions are imposed afterwards. 
\item The formula $\texttt{lane-1} \ U \ \texttt{behind}$ (read ``Lane~$1$ until behind'') specifies traces in which the agent drives continuously in Lane~$1$ until it is behind some blue car. No restrictions are imposed afterwards. 
\item The formula $F \ \texttt{lane-1}$ (read ``eventually Lane~$1$'') specifies traces in which the agent visits Lane~$1$ at least once, e.g., traces that end with the green car in Lane~$1$.

\end{itemize}
\end{example}

Formally, the syntax of LTLf is defined recursively. Each $p \in P$ is an LTLf formula.  If $\varphi_1$ and $\varphi_2$ are LTLf formulas, then so are $\varphi_1 \wedge \varphi_2$, $\neg \varphi_1$, $X \varphi_1$ (read ``next $\varphi_1$''), and $\varphi_1 U \varphi_2$ (read ``$\varphi_1$ until $\varphi_2$''). We use the abbreviation $F \varphi$ (read ``eventually $\varphi$'') for the formula $\texttt{True} U \varphi$.


The semantics of LTLf is defined by induction on the structure of the formula. Consider an LTLf formula $\varphi$ over $P$ and an abstract trace $\eta = \sigma_1, \ldots, \sigma_k$, where  $\sigma_i \in 2^P$, for $1 \leq i \leq k$. We say that $\eta$ satisfies $\varphi$, denoted $\eta \models \varphi$, when:

\begin{itemize}
\item If $\varphi =p \in P$, then $\eta \models \varphi$ iff $p \in \sigma_1$.
\item If $\varphi = \varphi_1 \wedge \varphi_1$ then $\eta \models \varphi$ iff $\eta \models \varphi_1$ and $\eta \models \varphi_2$.
\item If $\varphi = \neg \varphi_1$ then $\eta \models \varphi$ iff $\eta \not \models \varphi_1$.
\item If $\varphi = X \varphi_1$ then $\eta \models \varphi$ iff $(\sigma_2,\ldots,\sigma_k) \models \varphi_1$.
\item If $\varphi = \varphi_1 U \varphi_2$ then $\eta \models \varphi$ iff there is an index $1 \leq i \leq k$ such that $(\sigma_i,\ldots,\sigma_k) \models \varphi_2$ and for each $1 \leq j \leq i$, we have $(\sigma_j,\ldots, \sigma_k) \models \varphi_1$.
\end{itemize}

\paragraph{Nondeterministic finite automata}
Our algorithm to process queries is based on automata. A deterministic automaton (DFA, for short) is a tuple $\A = \zug{\Sigma, Q, \delta, q_0, Acc}$, where $\Sigma$ is an alphabet, $Q$ is a set of states, $\delta: Q \times \Sigma \rightarrow Q$ is a transition function, $q_0 \in Q$ is an initial state, and $Acc \subseteq Q$ is a set of accepting states. A run of $\A$ on a word $w = \sigma_1 \sigma_2 \ldots \sigma_k$, where $\sigma_j \in \Sigma$, for $1 \leq j \leq k$, is $r = r_0, r_1,\ldots, r_k$, where $r_i \in Q$, for $0 \leq i \leq k$, where $r$ starts in an initial state, i.e., $r_0 = q_0$, and respects the transition function, i.e., for each $i \geq 1$, we have $r_i \in \delta(r_{i-1}, \sigma_i)$. We say that $r$ is {\em accepting} if it ends in an accepting state, i.e., $r_k \in Acc$, and that $\A$ {\em accepts} $w$ if there is an accepting run on $w$. The {\em language} of $\A$, denoted $L(\A)$, is the set of words that it accepts. 

\begin{theorem}
\label{thm:LTLf}
\cite{LTLf}
Consider an LTLf formula $\varphi$ over a set of predicates $P$. There is a DFA $\A_\varphi$ over the alphabet $\Sigma = 2^P$ whose language is the set of traces that $\varphi$ recognizes. That is, for every trace $\eta \in \Sigma^*$ we have $\eta \in L(\A)$ iff $\eta \models \varphi$. 
\end{theorem}

\subsubsection{A Logic for Expressing Queries}
\label{sec:queryLang}
ASQ-IT is intended for laypeople in logic. That is, we do not assume that its users are capable of producing LTLf queries. In order to make ASQ-IT accessible, we develop a restricted query language, which is a fragment of LTLf. The query interface is built so that each query provided by a user maps to a formula in our language. We designed our language to be both accessible and expressive enough based on pilot studies so that users are capable of expressing properties of interest. We provide experimental evidence of its usability and effectiveness (see Sections~\ref{sec:usability}). Developing accessible fragments of logics is common practice in verification (e.g.,~\cite{dwyer1998property,de2014reasoning,berger2019multiple}). 

Let $P$ be a set of predicates. A {\em query}  is based on the following components: 
\begin{itemize}
\item A description of the start and end state of the trace. These are given as propositional formulas $\phi_s$ and $\phi_e$ over the predicates $P$. For example, when $\phi_s = \neg \texttt{lane-1} \wedge \texttt{behind}$, in any trace returned to the user, in the first position of a trace the green car is not in Lane~$1$ and behind some car. 
\item A constraint on the trace between $\phi_s$ and $\phi_e$, which is given as a third propositional formula $\phi_c$ over $P$.
\end{itemize}

Below, we describe several constraints that we implemented in our query interface. 

\begin{itemize}

\item The constraint $\phi_c$ {\em changes} is written in LTLf as $(\phi_s \wedge \phi_c) \wedge X \ F (\neg \phi_c \wedge F \phi_e)$. For example, for $\phi_c = \texttt{lane-2}$ (depicted in the query interface in Fig.~\ref{fig: tool}), the query represents traces that start with the agent driving in Lane~$2$ and at some point in the trace, the agent changes lanes. 

\item The constraint $\phi_c$ {\em stays constant} is written in LTLf as $(\phi_s \wedge \phi_c) \wedge X (\phi_c U \phi_e)$. For example, for $\phi_s = \texttt{lane-1}\wedge \texttt{behind}$, $\phi_e = \texttt{lane-4}$, and $\phi_c = \texttt{behind}$, the query represents traces that start with the agent driving in Lane~$1$ behind some car and ends when the agent is in Lane~$4$, and it drives behind some car throughout the whole trace.

\item The constraint $\phi_c$ {\em changes into} $\phi'_c$ is written in LTLf as $(\phi_s \wedge \phi_c \wedge \neg \phi'_c) \wedge X \ F (\neg \phi_c \wedge \phi'_c \wedge F \phi_e)$.
For example, for $\phi_c = \texttt{lane-1}$ and $\phi'_c = \texttt{lane-2}$, the query represents traces that start with the agent driving in Lane~$1$ and at some point switches to Lane~$2$. 
\end{itemize}

\subsubsection{Query Specification Interface}
We conducted  pilot studies to guide an iterative design process of the query specification interface, as well as the underlying LTLf fragment we chose to implement. This process resulted in the design of a simple interface using drop-down menus (see Figure~\ref{fig: tool}). The drop-down menu is designed to clearly and simply guide users toward possible state specifications for constructing their queries. Predicates, i.e. state attributes, are grouped into types to reduce cognitive load and avoid excessive options. For instance, all lane specifications appear under one drop-down, as these are mutually exclusive. 



\begin{remark}
\normalfont 
As we describe next, our backend is capable of processing general LTLf queries. Thus, it requires minimal effort to enhance the expressivity of the query interface as long as queries are mapped to LTLf. For example, previous versions of our tool allowed specifying intermediate states, e.g., a user might be interested to view a ``zig zag'' behavior: traces that start from Lane~$1$, visit Lane~$4$, and end in Lane~$1$. In LTLf, such behavior is specified as a concatenation of queries as described above. 
\end{remark}

\subsection{Backend: Processing User Queries}
\label{sec:algorithm}
Recall that offline, we collect a trace $s_1, \ldots, s_n$ of the agent operating in its environment, and a domain expert provides a collection of predicates $P$ with which we obtain an abstract trace $P(s_1),\ldots, P(s_n)$. In addition, we assume that a user provides an LTLf query $\varphi$. We describe an algorithm to process the user's query, formally stated as follows.

\noindent{\bf Problem:} Given an LTLf query $\varphi$, find a sub-trace $P(s_k),\ldots, P(s_\ell)$ that satisfies $\varphi$.

\paragraph{The algorithm}
Consider a trace $\eta = P(s_1),\ldots, P(s_n)$ over $2^P$ and an LTLf formula $\varphi$. We construct two DFAs $\A_\varphi$ and $\A_{F \varphi}$ (read ``eventually $\varphi$'') as in Thm.~\ref{thm:LTLf}. Note that $\A_{F \varphi}$ accepts all traces that end in a suffix that satisfies the user's query $\varphi$. We feed $\eta$, letter by letter to $\A_{F \varphi}$ until it visits an accepting state. Suppose that $\eta' = P(s_1),\ldots, P(s_\ell)$ is accepted by $\A_{F \varphi}$, then we are guaranteed that $\eta'$ has a suffix that satisfies $\varphi$. 

Next, we search for the beginning of the suffix, i.e., an index $k < \ell$ such that $P(s_k), \ldots, P(s_\ell)$ satisfies $\varphi$. We read the trace backward, starting from $P(s_\ell)$ and until $P(s_1)$ while executing ${\A}_\varphi$ ``backward'', starting from the accepting states of $\A_\varphi$ and until an initial state is visited. Formally, we maintain a set of states $Q' \subseteq Q$, which is initialized to $Acc$. When reading a letter $\sigma$, we update $Q'$ to be $\set{q \in Q: \exists q' \in Q' \text{ s.t. } q' = \delta(q, \sigma)}$. We terminate once $q_0 \in Q'$. It is not hard to show that if the algorithm terminates after $P(s_k)$ is read, then the suffix $P(s_k),\ldots, P(s_\ell)$ satisfies $\varphi$. Moreover, note that $\eta$ is read (forward) once by $\A_{F \varphi}$ and read at most once (backward) by $\A_\varphi$, thus the running time is linear in $n$. 

\begin{theorem}
Consider a collection of predicates $P$ and a trace $\eta$ over $2^P$ of length $n$. Given an LTLf formula $\varphi$, the algorithm returns a sub-trace that satisfies $\varphi$, if one exists. The algorithm processes $\eta$ at most twice.
\end{theorem}

\begin{remark}
\normalfont 
Once the algorithm finds a trace $P(s_k),\ldots,P(s_\ell)$ that satisfies $\varphi$ it restarts from index $\ell+1$ in search for another query until reaching the end of the database. In our implementation, a query might be answered by numerous clips, dependent on the database size. 
\end{remark}

\section{Empirical Evaluation}
To evaluate ASQ-IT, we conducted two user studies. The first was a usability study with laypeople who have no prior knowledge of AI or reinforcement learning, to examine their ability to understand and formulate queries in ASQ-IT. The second study assessed ASQ-IT's benefits for users with some AI knowledge who may use such a tool for debugging. 


\subsection{User Study 1: Usability Assessment}
\label{sec:usability}
The goal of this study is to examine laypeople's interaction with ASQ-IT and test its usefulness and effectiveness for generating queries to an agent.


\subsubsection{Empirical Methodology}
\emph{Agent.} We trained a policy for 2000 episodes using a double DQN architecture and penalized for collisions.

\emph{Participants.} Forty participants were recruited through Prolific (20 female, mean age $= 34.7$, STD $= 11.29$), each receiving $\$4.5$ for their completion of the Task. To incentivize
participants to make an effort, they were provided a bonus of 15 cents for each
correct answer. Participants whose overall task duration was lower than the mean by more than two standard deviations were filtered out. 

\emph{Procedure.}
First, participants were introduced to the Highway domain and the concept of AI agents. Then commenced an introduction to the ASQ-IT's interface and the process of generating queries for the system. Each explanation was followed by a short quiz to ensure understanding before advancing. As a final step before the task, participants were provided a link to ASQ-IT's interface where they could interact and explore both the interface and the agent. 
In the task section, participants were tested on their understanding of the interface, query generation, and output through three types of tasks. \emph{i) Movies to Queries (\textbf{M2Q}):} Given an output video, select the correct query that would result in its generation (example in supplementary).
\emph{ii) Free Text to Queries (\textbf{T2Q}):} Given textual descriptions of desired behavior, select the correct query.
\emph{iii) Queries to Free Text (\textbf{Q2T}):} Given a query, select the correct textual description of the desired behavior.   
All questions were multiple-choice with four possibilities and a single correct answer and each task type included two questions in ascending difficulty \footnote{Full user study available at \url{https://bit.ly/3GJV394}}.
Upon task completion, participants were prompted to provide textual feedback regarding their experience with the system \& interface and complete a usability survey~\cite{brooke1996sus}.

\subsubsection{Results \& Discussion}
We analyzed participants' responses in terms of objective performance, usability ratings, and textual responses. The quantitative results are summarized in Figure~\ref{fig:study2results} (A,B,C). We discuss the main findings and insights based on users' responses.

\emph{Participants were able to comprehend the semantics of our logic \& use it to formulate meaningful queries.}
Overall, participants were successful in the tasks of interpreting queries and formulating queries (Figure~\ref{fig:study2results}A). In all questions, participants did significantly better than a random guess, and in 4 out of 6 questions success rate was $\approx 90\%$. 
We find these results highly encouraging: ASQ-IT allows participants with no training in logic to express behavior as formal queries in LTLf and to understand their output.

We identified two main causes for incorrect  answers: (1) \textit{Agent relations (position):} Confusing the position of the agent compared to other cars such as mixing ``Behind'' with ``In Front Of'' (e.g. is the agent behind another car or is there one behind the agent?), and (2) \textit{Misunderstanding constraints:} Some participants were not able to understand the use of constraints on the agent's trace and most often chose to ignore these specifications.
These alone were responsible for $\approx 90\%$ of all incorrect answers. 
\emph{Participants improved throughout the task.} Some participants who struggled with simple questions regarding constraints would manage to solve correctly harder questions that appeared later. Some participants noted that elements of the interface became clearer when asked to answer questions about them. One participant wrote 
\textit{``I found the instructions quite hard to understand. When a description was provided and you had to complete what you thought was the correct specification, I found this a better way to learn the process.''}


\emph{Participant reported high effectiveness scores.}
Following \citeauthor{brooke1996sus}'s\shortcite{brooke1996sus} system usability scale, participants found ASQ-IT, on average, more effective than not, in all categories (see Figure~\ref{fig:study2results}B). Effectiveness is the measurement of a tool's ability to produce the desired outcome. Multiple responses mentioned its usefulness for testing and observing how the agent acts. Others described positively the fact that it was clear to them what videos would be generated by ASQ-IT, so long as the specification was not very complex, and after some initial trial and error phase.
Most negative responses mentioned the many options available and the complexity of understanding the interface. However, many participants reported that after some exploration, their experience and understanding greatly improved, suggesting a learning curve in using the tool. 

\emph{Participants reported an increase in efficiency over tool usage.} Efficiency measures the ease of using a tool. Many participants described some level of uncertainty upon initial interaction with the interface, mainly given the lengthy explanations prior to using it. However, the majority of participants reported quickly understanding once access to ASQ-IT was given and some exploration of the interface was conducted. When asked what would help them interact with the tool, many participants responded that they would prefer the interface to have fewer options and more visual aid for the existing ones. 

\emph{Expressivity.}
When asked to describe what features or behaviors were missing or desired for the highway domain, participants mostly requested the ability to control the agent's speed and distance from other cars, along with the option to specify the positions of other cars and the output video length.
When asked what agent behaviors and situations were of interest to them, specifiable or not using ASQ-IT's current interface, participants mostly referred to observing the agent react to critical situations such as obstacles on the road, lane merges or interaction with other cars such as emergency vehicles or evasion of accelerating or braking cars.

\subsection{User Study 2: Identifying Agent Faults}
We conducted a second user study to assess how users interact with ASQ-IT when working on a task and whether using ASQ-IT improves their performance. To this end, we simulated faulty agents and tested participants' ability to debug them through exploration and investigation. The study had two main goals: (1) to understand the process of querying agent behavior using ASQ-IT, and (2) to assess the usefulness of ASQ-IT in a debugging task compared to a static policy summary explanation method 
~\cite{amir18highlights}. 


\subsubsection{Empirical Methodology}
\begin{figure}
    \frame{\includegraphics[width=0.7\columnwidth]{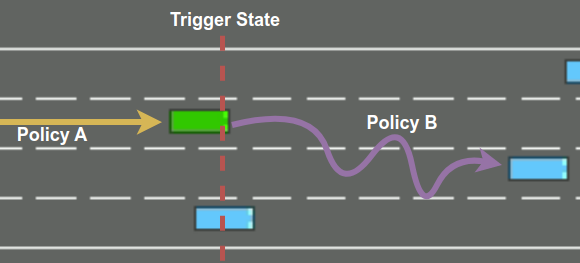}}\\
    \caption{Simulating a faulty agent. The \textit{Plain-Collision} setup.}
	\label{fig: faulty agents}
	\vspace{-0.2cm}
\end{figure}

\emph{Agents.} 
We trained three agents for 2000 episodes using the double DQN architecture. To simulate a faulty agent, we combined two of the agents $Agt_1$ and $Agt_2$ into one. We choose a {\em trigger} event, e.g., the agent is on Lane~$2$ and below a car. Initially, $Agt_1$ operates, and once the trigger event occurs, control is passed to $Agt_2$ (see Fig.~\ref{fig: faulty agents}). Specifically, we used (1) \textit{\underline{Plain-TopLane}}: $Agt_1$ is a simple agent used in  the usability study, and $Agt_2$ prioritizes the top-most lane, and (2) \textit{\underline{Plain-Collision}}: $Agt_1$ is the same simple agent and $Agt_2$ tries to collide with other cars. 


\emph{Participants.} Since we used a fairly complex debugging task, our target users were people who have some knowledge of AI. We recruited thirteen participants from the university who have completed at least one AI or machine learning course (2 female, mean age $= 29.3$, STD $= 5.1$). Participants received \$$15$ for their participation. The experiment took on average 45 minutes to complete. 

\emph{Conditions.} Participants were assigned to either the ASQ-IT system or a system that implemented the HIGHLIGHTS policy summarization algorithm~\cite{amir18highlights}. We intentionally recruited more participants for the ASQ-IT condition (8 for ASQ-IT, 5 for HIGHLIGHTS), as we were interested in learning about the interaction with the system. 
Participants in the ASQ-IT condition interacted with the system through queries which they could construct using drop-down menus (Figure~\ref{fig: tool} - Query Interface). Submitting a query would provide participants with up to four videos which answer the query, chosen randomly from the set of all such videos. An option to load more videos was available given that more such videos existed.
Participants in the HIGHLIGHS condition were presented with a simple interface that only included a single video and an option to load the next video or return to the previous one. Forty videos were made available this way, appearing in a sorted fashion based on the importance assigned to them by the HIGHLIGHTS algorithm. We made use of the HIGHLIGHTS-DIV variant of the algorithm that also takes into consideration the diversity between videos such that the videos were unique and captured multiple important states and not solely the most important one.
We made sure both condition videos were of similar parameters such as FPS and minimum length.

\emph{Tasks.}
The study consisted of three main tasks: (1) elimination, (2) hypothesis generation, and (3) verification. {\underline{\textit{Elimination}}: Participants explored the \textit{Plain-TopLane} faulty agent using their assigned explanation system. Participants were required to identify the correct trigger from a list of four options and to describe in free text the behavioral change that occurs following the trigger event. \underline{\textit{Hypothesis generation}}: Participants were shown two videos of the \textit{Plain-Collision} faulty agent in which the trigger event and the behavior change appear. They were told what the fault was (i.e., trying to collide with other cars) and were asked to hypothesize what trigger event caused the change in behavior. Participants were also asked to describe how they would use the system to refute or validate their hypothesis. \underline{\textit{Verification}}: was to try to refute or validate their proposed hypotheses using the explanation system and, if need be, raise new ones.

\emph{Procedure.}
First, participants were introduced to the Highway domain and its key elements. Next, participants were familiarized with the explanation system they would be using, either the ASQ-IT interface, or an interface for watching HIGHLIGHT videos. During this instructions phase, participants could interact with the system and understand how to work it (this was optional, but all participants chose to do so). When satisfied, participants moved on to the study tasks.  All tasks included a confidence rating question on a 1 to 7 Likert scale. Lastly, participants answered an explanation satisfaction survey based on \cite{hoffman2018metrics}, provided textual feedback on the system they used, and answered demographic questions\footnote{Full user-study available at \url{https://bit.ly/3GJV394}}. All sessions were done in the presence of the first author who encouraged participants to think aloud. The sessions were recorded, including both the computer screen and the audio. Participants' actions in the system were logged.

We assigned success scores to participants based on the relation between their answer and the correct trigger event: 
(1) No relation: 0 points, (2) Partial relation (e.g, specifying only one of two conditions for the trigger event): 1 point, (3) Exact trigger included (when multiple hypotheses raised): 2 points, and (4) Exact trigger chosen: 3 points. 

\begin{figure}
  \includegraphics[width=1\linewidth]{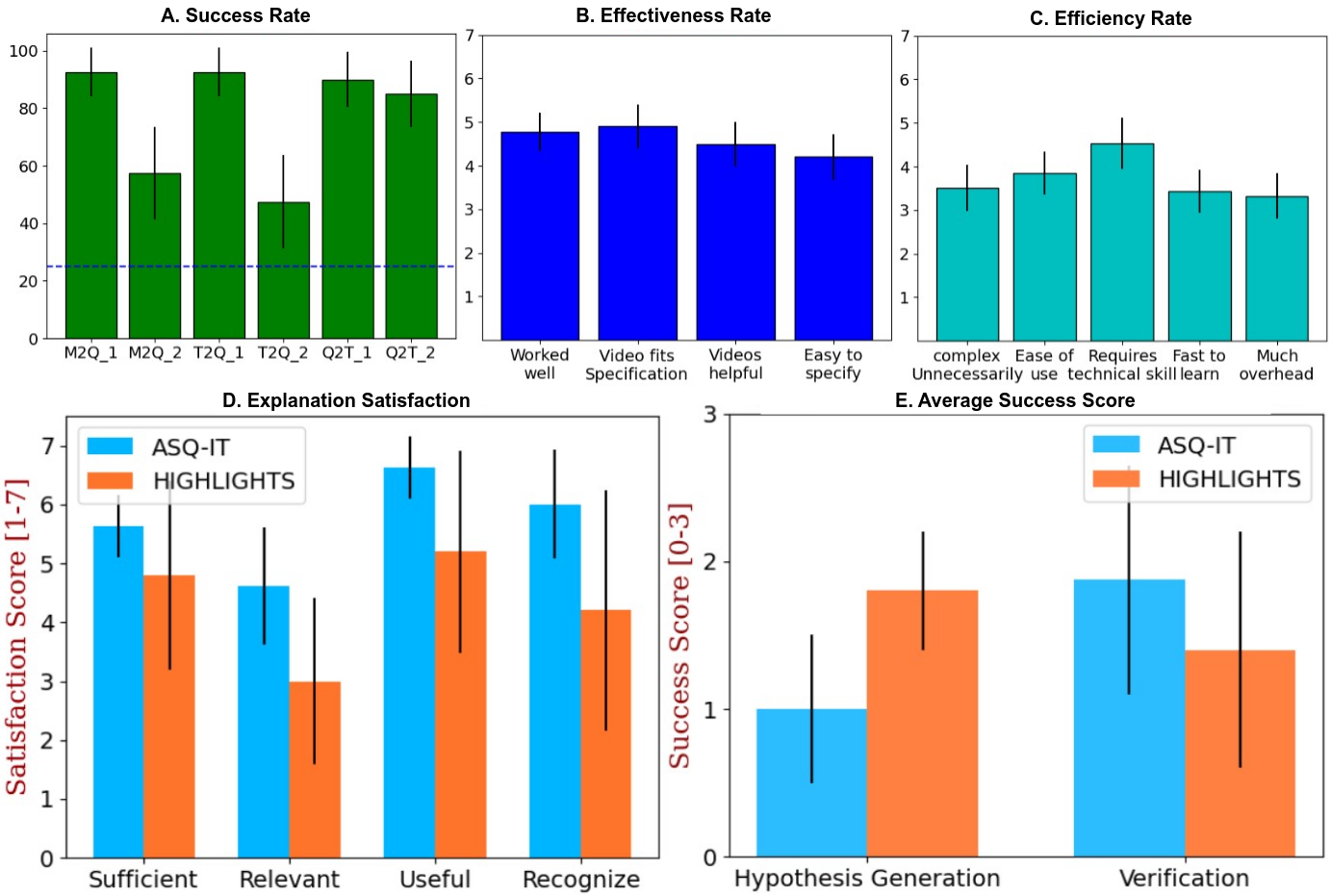}
  \caption{ \textbf{Top:} Usability Study Results. \\ \textbf{Bottom:}Agent Faults Study Results.}
  \label{fig:study2results}
  \vspace{-0.4cm}
\end{figure}

\subsubsection{Results \& Discussion}
We report the main observations regarding participants' experience and performance with ASQ-IT, and compare it to the use of HIGHLIGHTS. 
We analyzed participants' activities based on the session recordings and system logs. 
We report the average scores of participants as well as the average explanation satisfaction ratings in Figure~\ref{fig:study2results} (D,E). As we are mainly interested in the process of using different explanation approaches, we elaborate on qualitative observations made based on the analysis of participants' activities.



\emph{ASQ-IT participants revised their hypotheses.}
Six out of eight ASQ-IT participants revised the hypotheses they generated in the second task based on the explanation videos outputted by their queries to the ASQ-IT interface, while the other two were confident in theirs and chose to keep them.
Meanwhile, only one participant in the HIGHLIGHTS condition made a revision to their original hypothesis. 

\emph{Most ASQ-IT participants who revised their hypotheses improved their identification of the trigger event.}
Out of the six ASQ-IT participants who revised their hypotheses, four were able to improve their score on the final answer. The remaining two participants maintained the same score. In contrast, the HIGHLIGHTS participant who revised her initial hypothesis received a lower score for her final answer compared to her initial response. The average change in participant success is illustrated in Figure~\ref{fig:study2results}E.

\emph{Participants' method of hypothesis verification differed significantly between conditions.}
This was most evident in the \emph{elimination} task.
ASQ-IT participants were able to choose which trigger to inspect, define it as a query and observe videos of the agent in these situations. They were all able to eliminate options until reaching the correct answer. For six out of eight participants, the correct trigger became immediately evident once queried. The two remaining participants required additional queries in order to be convinced before ultimately selecting the correct trigger. Apart from one participant, who struggled initially with the interface, mostly due to confusion regarding the role of the constraint drop-downs, all other ASQ-IT participants resolved the elimination task quickly and reported it as easy.     

HIGHLIGHTS participants, on the other hand, had no control over the videos they received, and as such were forced to see each movie without knowing which trigger option might appear. Four out of five participants' process involved associating each movie with a possible trigger in the list, while the remaining participant searched videos for noticeable patterns and then compared them to the list. Both processes become both tedious and cognitively overwhelming as the number of options grows, especially when there is no guarantee that any of the trigger options will appear. In referral to their decision process for the final answer, all noted that the task was not easy and that their answers are mostly based on which triggers they have seen most in the videos. 

\emph{ASQ-IT participants who identified the correct trigger were able to verify it.}
Out of five ASQ-IT participants that identified the correct trigger (at some point), four were able to verify it using ASQ-IT and submit the correct answer. A typical verification process involved formulating queries that specified  hypothesized trigger events and reviewing the retrieved video clips to see whether these indeed led to the behavior change.
Meanwhile, three out of five HIGHLIGHTS participants refuted the correct hypothesis in favor of a more general, but partial answer. This can be associated with the same loss of confidence derived from self-reported lack of control over explanation videos as further discussed below. 

\emph{ASQ-IT participants calibrated their confidence.}
Six out of eight ASQ-IT participants adjusted their reported confidence in a justifiable way based on their interaction with the system. These include two participants that adjusted upwards due to successfully identifying the correct trigger and four participants adjusting downwards based on either the need for revisions or the recognition that the exact answer was not found.
The remaining two participants either experienced no confidence change due to recognizing the correct trigger and validating it or were unaware of their partial solution due to confirmation bias which raised their confidence needlessly. 
While interesting, we take this observation with a grain of salt as there are typically substantial individual differences in confidence and the sample size is small.

Four out of five HIGHLIGHTS participants also calibrated their confidence. Three of them lowered their confidence and commented that they were not able to view the videos that they thought would help them validate or refute their hypothesis. That is, in contrast to the ASQ-IT participants who lowered their confidence due to observing information that did not align with their hypothesis, HIGHLIGHTS participants lowered their confidence because the system did not provide them with helpful information.
\emph{ASQ-IT participants reported higher explanation satisfaction.}
Upon completion of study tasks, HIGHLIGHTS participants reported more frustration and less satisfaction regarding the explanation system they were assigned, as can be seen in Figure~\ref{fig:study2results}D. All HIGHLIGHTS participants mentioned feeling a lack of control regarding the videos they were shown, four participants stated difficulty in validating or refuting their hypotheses, and three reported loss of confidence. 
One of the participants summarized these difficulties in a concise manner, stating that \textit{``Lack of variance [in videos] ... hard to refute hypotheses''} and \textit{``Lack of consistency [in videos] ... hard to validate hypotheses''}.
ASQ-IT participants, on the other hand, largely reported a very positive experience with the explanation system. 
This positive experience was also evident both in participants' feedback section where they suggested features and options they'd like the system to allow in the future, and vocally off-record upon experiment termination.

\section{Summary and Future Work}
We  developed ASQ-IT -- an XRL interactive tool for querying AI agents that utilizes formal 
methods. Results from two user studies demonstrate that the tool is usable even to laypeople and that it supported users with no background in temporal logic
in an agent-debugging task. In the debugging task, the tool was more useful than a baseline static explanation approach, as it enabled users to specify the information that they wish to explore regarding the agent's policy. Beyond the improvement in participants' objective performance in the task, there were noticeable differences in the process of exploring agent behavior. In particular, participants using ASQ-IT were more engaged, open to new hypotheses and felt more in control compared to participants using the static explanation. These findings highlight the potential benefits of designing more interactive explanation methods.

There are several directions that can be explored in future work.
A key question in the design of the tool is the balance between expressivity and complexity. It is possible that alternative interface designs could provide better scaffolding for more complex queries, such that users could gradually extend their ability to examine policies. Moreover, it would be interesting to go beyond the specifications of state predicates and develop a language for describing more abstract queries about the behavior of the agent 
(e.g., allowing users to query for ``risky'' behaviors). 
The tool could also be improved by integrating into it a variety of existing explanation methods, such that users could alternate between different pre-specified explanations and specifying their own queries. Such pre-specified explanations may help users identify which aspects of the agent's policy should be explored further.

\bibliographystyle{named}
\bibliography{bib}

\end{document}